\documentclass[sigconf]{acmart}

\usepackage{amsmath,bm}
\usepackage{booktabs}
\usepackage{multirow}
\usepackage{makecell}
\usepackage{color}
\usepackage{enumitem}
\usepackage{algorithm}
\usepackage{algorithmic}
\usepackage{CJKutf8}
\usepackage{diagbox}
\usepackage{graphicx}
\usepackage{graphicx}
\usepackage{subfigure}
\usepackage[normalem]{ulem}
\useunder{\uline}{\ul}{}

\AtBeginDocument{
  }

\setcopyright{acmlicensed}
\copyrightyear{2024}
\acmYear{2024}
\acmConference[WSDM '24]{the seventeenth ACM International Conference on Web Search and Data Mining}{March 4--8, 2024}{Mérida, Yucatán, Mexico}
\acmBooktitle{the seventeenth ACM International Conference on Web Search and Data Mining (WSDM '24), March 4--8, 2024, Mérida, Yucatán, Mexico}
\acmDOI{XXXXXXX.XXXXXXX}

\begin{document}
\title{RDGCN: Reinforced Dependency Graph Convolutional Network for Aspect-based Sentiment Analysis}

\author{Xusheng Zhao}
\affiliation{
  \institution{Institute of Information Engineering,\\ Chinese Academy of Sciences $\And$}
  \institution{School of Cyber Security, University\\ of Chinese Academy of Sciences}
  \country{}
}
\email{zhaoxusheng@iie.ac.cn}

\author{Hao Peng}
\affiliation{
  \institution{School of Cyber Science and Technology,\\ Beihang University}
  \country{}
}
\email{penghao@buaa.edu.cn}

\author{Qiong Dai}
\authornote{Corresponding author.}
\author{Xu Bai, Huailiang Peng}
\affiliation{
  \institution{Institute of Information Engineering,\\ Chinese Academy of Sciences}
  \country{}
}
\email{{daiqiong,baixu,penghuailiang}@iie.ac.cn}

\author{Yanbing	Liu}
\affiliation{
  \institution{Institute of Information Engineering,\\ Chinese Academy of Sciences}
  \country{}
}
\email{liuyanbing@iie.ac.cn}

\author{Qinglang Guo}
\affiliation{
  \institution{University of Science and Technology of China}
  \country{}
}
\email{gql1993@mail.ustc.edu.cn}

\author{Philip S. Yu}
\affiliation{
  \institution{Department of Computer Science,\\
  University of Illinois Chicago}
  \country{}
}
\email{psyu@uic.edu}

\renewcommand{\shortauthors}{Zhao et al.}

\begin{abstract}
Aspect-based sentiment analysis (ABSA) is dedicated to forecasting the sentiment polarity of aspect terms within sentences.
Employing graph neural networks to capture structural patterns from syntactic dependency parsing has been confirmed as an effective approach for boosting ABSA.
In most works, the topology of dependency trees or dependency-based attention coefficients is often loosely regarded as edges between aspects and opinions, which can result in insufficient and ambiguous syntactic utilization.
To address these problems, we propose a new reinforced dependency graph convolutional network (RDGCN) that improves the importance calculation of dependencies in both distance and type views.
Initially, we propose an importance calculation criterion for the minimum distances over dependency trees.
Under the criterion, we design a distance-importance function that leverages reinforcement learning for weight distribution search and dissimilarity control.
Since dependency types often do not have explicit syntax like tree distances, we use global attention and mask mechanisms to design type-importance functions.
Finally, we merge these weights and implement feature aggregation and classification.
Comprehensive experiments on three popular datasets demonstrate the effectiveness of the criterion and importance functions.
RDGCN outperforms state-of-the-art GNN-based baselines in all validations.
\end{abstract}



\maketitle

\vspace{-2mm}
\section{Introduction}
\label{sec_1}
\begin{figure}[t]
\centering
\includegraphics[width=\columnwidth]{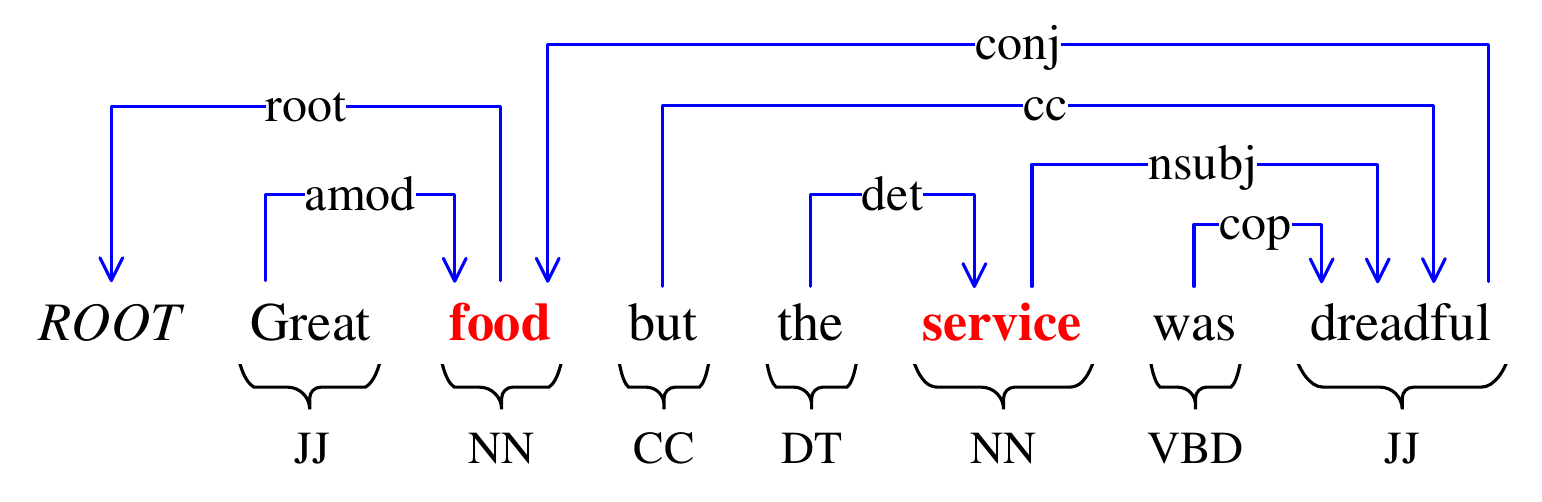}
\vspace{-9mm}
\caption{
An example sentence and its dependency tree, with aspect terms shown in bold red.
{\it ``ROOT''} is a virtual word, and the symbols below each real word represent parts of speech.
}
\label{Fig_Intro}
\vspace{-2mm}
\end{figure}

Aspect-based sentiment analysis (ABSA) is a fine-grained task that focuses on predicting the sentiment polarity of aspect terms within sentences~\cite{zhang2022survey}.
The sentence {\it ``Great food but the service was dreadful''} in Figure~\ref{Fig_Intro} serves as an example, with the aspects {\it ``food''} and {\it ``service''} exhibiting positive and negative sentiments, respectively.

Early ABSA research~\cite{schouten2015survey} primarily relied on manually designed syntactic features.
A large number of neural network methods have recently emerged~\cite{do2019deep}, which are non-labor-intensive and bring huge performance improvements.
Since context words in a sentence may have different importance for a given aspect, attention mechanisms are widely integrated in recurrent neural networks~\cite{wang2016attention,ma2017interactive}, memory networks~\cite{tang2016aspect}, gated networks~\cite{zhang2016gated, xue2018aspect}, and convolutional networks \cite{xue2018aspect}, etc.
However, there may be multiple aspect terms and different opinions in a sentence.
Attention-based weighting can make aspect representations susceptible to interference from irrelevant opinions.
Taking Figure~\ref{Fig_Intro} as an example, for the aspect {\it ``service''}, both opinion words {\it ``dreadful''} and {\it ``Great''} may be assigned large attention scores, with the latter potentially hindering sentiment assessment.

Thanks to advances in syntactic parsing with neural networks, the dependency trees of sentences are becoming more accurate~\cite{dai2021does}, prompting studies, such as~\cite{zhang2019syntax,zhou2021closer,zhang2022survey}, to model explicit connections between aspects and their associated opinion words.
As a paradigm good at learning topological data, graph neural network (GNN)~\cite{wu2020comprehensive} is widely applied in ABSA methods to utilize dependency structures.
Based on the syntax within dependency trees, GNN-based methods are typically divided into three streams.
The first branch exploits the discrete (or probabilistic) topology of trees \cite{zhang2019aspect,sun2019aspect,huang2019syntax,tang2020dependency,liang2020jointly,hou2021graph,li2021dual}.
The second branch focuses on the diversity of dependency types in trees \cite{zhang2020convolution,wang2020relational,tian2021aspect,tang2022affective}.
The third branch utilizes tree-based minimum distances, that is, the number of edges on the shortest path between two words~\cite{wang2020relational,zhang2022ssegcn,zhao2022graph}.
Regularly, type- and distance-based methods involve using the raw topology of dependency trees.
This is because types are attached to the topology, which in turn is the special case of minimum distances all equal to 1~\cite{chen2022enhanced}.
Despite the success of these studies, issues of underutilization of syntax still persist.

First, because the syntax of types and distances is fundamentally different, intuitively applying the same processing strategy to them may be insufficient~\cite{wang2020relational,chen2022enhanced}.
For example, for dependency types, since their weights are implicit without experts, it is workable to calculate the importance based on the attention mechanism~\cite{wang2020relational,tian2021aspect}.
However, the attention coefficients may obscure the original explicit syntactic importance of the minimum tree distances, resulting in meaningless calculations from scratch~\cite{wang2020relational}.
Second, how to effectively implement calculations for explicit distance weights is underexplored, which is not limited to GNN-based research.
Most studies, such as \cite{zhang2019syntax,phan2020modelling,zhao2022graph}, equidistantly down weights for induced dependencies in ascending order of tree-based distances.
Even though this strategy is proven to be effective in preserving and distinguishing distance syntax, it still suffers from an unreasonable equidistant setting.
For example, for sentences with a large distance range, usually, only a small number of dependencies with small distances need to be distinguished by importance.
In contrast, the importance of dependencies with too large distances may all be close to 0 rather than equidistant.

To address these problems, we propose a reinforced dependency graph convolutional network (RDGCN) using different importance calculations for dependency types and distances.
Specifically, at first, we propose a new importance calculation criterion for the minimum distances over dependency trees, which imposes constraints on both importance discriminability and distribution.
Second, according to this criterion, we propose a distance-importance function consisting of two sub-functions.
More specifically, to increase discriminability, a power-based sub-function sets the $[0,1]$-weights of dependencies with the minimum and maximum distances to 1 and 0, respectively.
To avoid unreasonable arithmetic distribution, an exponential-based sub-function is designed. 
It disproportionately reduces the weights of induced dependencies, whose gradients gradually tend to parallel along the direction of increasing distance.
Given the absence of prior knowledge on the range of valuable distances, we use reinforcement learning (RL) \cite{vermorel2005multi} to search for the exponential curvature to control the concave arc distribution of the dependency weights.
In this way, the tree-based distance importance weights fall off smoothly rather than abruptly, preserving the possibility of exploiting dependencies with large distances.
Besides, RL-based automatic search for the best curvature avoids tedious manual parameter adjustment, rendering RDGCN highly portable across different ABSA tasks.
Third, because dependency types do not provide explicit syntactic importance such as minimum distances, we introduce a global attention mechanism to differentiate type weights.
Finally, we combine distance and type weights, and perform feature aggregation and sentiment prediction based on GCN.
The major contributions are summarized as follows:

$\bullet$ We propose a novel ABSA model that efficiently captures both distance and type syntax through different strategies.

$\bullet$ This work is an important attempt on how to calculate the non-equidistant importance for explicit distance syntax.

$\bullet$ We evaluate the proposed RDGCN on three popular datasets whose experimental results and analysis verify the rationality of the criterion and functions as well as the superiority of the performance.

\section{Preliminaries}
\label{sec_2}
In this section, we describe aspect-based sentiment analysis (ABSA) as well as graph neural network (GNN)-based ABSA.

\subsection{ABSA}
\label{sec_2_1}
Given a sentence-aspect pair $X-Y$, where $Y=<$$y_1, y_2, ..., y_M$$>$ is an aspect and sub-sequence of the sentence $X=<$$x_1, x_2, ..., x_N$$>$, $x_{i}$ and $y_{j}$ are the $i$-th and $j$-th words (tokens) in $X$ and $Y$, $N$ and $M$ are the lengths of the sentence and aspect, respectively. 
For ABSA, it needs to predict the sentiment polarity of $Y$ by drawing information from $X$, i.e., $X-Y\rightarrow C$, where $C$ is the polarity category like positive and negative.
The current dominant approach is to derive the contextual representations of sentences based on encoders such as Transformer \cite{vaswani2017attention} and BERT~\cite{kenton2019bert}.
The sequence $X$ can be transformed into a low-dimensional embedding matrix $\mathbf{E}\in\mathbb{R}^{N\times D}$, where the $i$-th row of $\mathbf{E}$ represents the feature vector $e_{i}$ with dimension $D$ of the $i$-th token.
Afterwards, aspect-specific features $\mathbf{F}\in\mathbb{R}^{M\times D}$ are derived from $\mathbf{E}$.

\begin{figure*}[t]
\centering
\includegraphics[width=\textwidth]{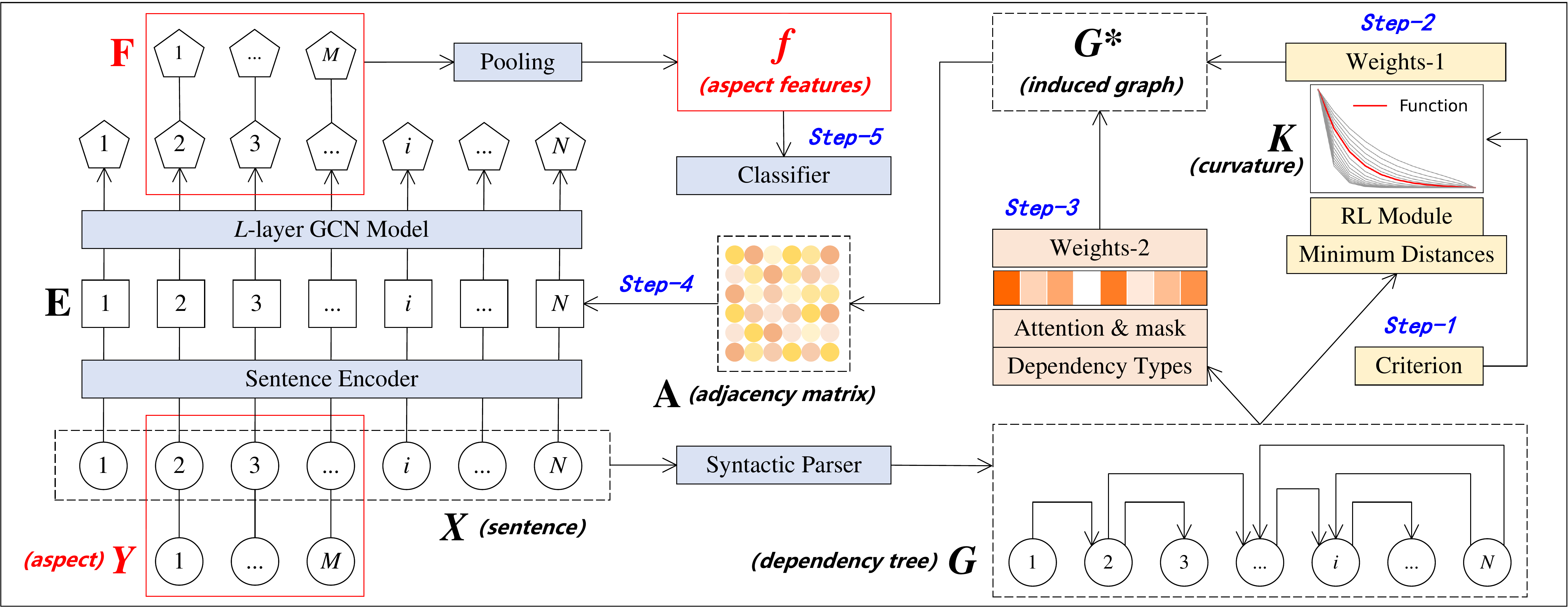}
\vspace{-6.5mm}
\caption{The overview of RDGCN, which comprises five steps: 
{\it \textbf{1) Criterion Construction}} for importance calculation of minimum distance syntax, 
{\it\textbf{2) Distance-importance Function}} designed according to the novel criterion, 
{\it\textbf{3) Type-importance Function}} designed for dependency type syntax,
{\it\textbf{4) Feature Aggregation}} on the induced syntactic graph to optimize token representations,
{\it\textbf{5) Pooling and Classification}} to obtain aspect-specific representations and sentiment polarities.}
\label{Fig_Framework}
\vspace{-1mm}
\end{figure*}

\subsection{GNN-based ABSA}
\label{sec_2_2}
GNN-based ABSA typically needs to introduce additional syntactic structures, which often come from dependency tree-based topology, type, and distance views.
For example, given a sentence-aspect pair $X-Y$, the dependency tree $G$ corresponding to the sentence $X$ can be yielded by an existing syntactic dependency parser.
Then, the graph $G$ can be abstracted as an adjacency matrix $\mathbf{A}\in\mathbb{R}^{N\times N}$, whose entry from the $i$-th row and $j$-th column indicates the dependency weight between the $i$-th and $j$-th tokens of $X$.
The dependency weights of $\mathbf{A}$ in different GNN-based models are usually obtained from different computational strategies.
Combined with the token feature matrix $\mathbf{E}$ output by the sentence encoder, GNN learns the syntactic structure patterns of sentences through feature aggregation.
The aggregation operation at the $l$-th layer can be formulated as the following form:
\begin{align}
	e^{l}_{i}=\sigma
    (e^{(l-1)}_{i}\oplus
	\text{AGG}^{l}(\{e^{(l-1)}_{j}:\mathbf{A}[i,j]>0 \})),
	\label{Eq_1}
\end{align}
where $e^{(l-1)}_{i}$ and $e^{l}_{i}$ are the input and output feature vectors of the $i$-th token (node) $x_{i}$.
$\mathbf{A}[i,j]>0$ represents that there is a dependency (edge) between tokens $x_{i}$ and $x_{j}$, and $e^{(l-1)}_{j}$ is the neighbor features of $x_{i}$ to be aggregated.
$\text{AGG}^{l}(\cdot)$ denotes an aggregation function like attention \cite{velivckovic2018graph} and convolutional \cite{kipf2017semi} operations, whose superscript $l$ usually denotes a specific in-layer feature transformation module.
Moreover, $\oplus$ is an operation to combine the features of $x_{i}$ and that of its neighbors, such as averaging and concatenation, and $\sigma(\cdot)$ is an activation function, such as Tanh and ReLU.
After the aggregation is completed at the final layer $L$ of GNN, the sentence feature matrix $\mathbf{E}^{L}\in\mathbb{R}^{N\times D}$ will be obtained and used for follow-up aspect-specific feature pooling and classification.
Overall, GNN-based ABSA has an additional sentence-graph conversion process, i.e., $X\rightarrow G-Y\rightarrow C$.

\section{Methodology}\label{sec_3}
In this section, we formulate the novel reinforced dependency graph convolutional network (RDGCN).
Its overview is shown in Figure ~\ref{Fig_Framework}.

\subsection{Criterion Construction}\label{sec_3_1}
Previous studies like \cite{zhang2019syntax,phan2020modelling,zhao2022graph,zhang2022ssegcn} introducing minimum distance syntax based on dependency trees, typically abide by the calculation criterion that the larger the distance, the less important the induced dependencies (edges) are.
Specifically, they usually decrease weights equidistantly from small to large distances based on reciprocals and differences.
Although such strategies are confirmed to preserve and differentiate distance syntax efficiently, the equidistant settings are often unrefined and unreasonable.
Considering that there is no work to explore how to effectively implement importance calculations for explicit distance syntax, we pioneer a universal calculation criterion.
More specifically, we argue that an excellent importance calculation criterion for minimum tree distances should impose requirements in terms of discriminability as well as distribution: 
{\it{\textbf{Discriminability} requires that the importance weights acquired from different distance values should be as discriminable as the distances themselves.}}
Because the importance gap between adjacent distance values is determined by the distribution, in this case, we only constrain the extreme of the weights.
In other words, the importance weights with the minimum tree distances of 0 (lower bound) and $T$ (upper bound) should be far apart, leaving space for differences in intermediate values.
Hence, for weights scaled to $[0,1]$, the maximum and minimum weights are best fixed at 1 and 0.
{\it{\textbf{Distribution} means that the weight distribution should be feasible and reasonable while ensuring discriminability.}}
For example, given a sentence with a very large upper bound $T$ of the minimum distances, since the syntax gradually blurs as the distance increases, the weight discrepancy between two adjacency distances with a larger value should be smaller than that with a smaller value.
In other words, the degree of discriminability of the importance of different distance intervals should not be the same.
Thus equidistant decreases that gives little attention on important distance intervals are sub-optimal.

Even though this criterion specifies the above two requirements, following it to design distance-importance functions still introduces three challenges.
Firstly, unlike the arithmetic sequence distribution of weights that requires only one linear function, there is no natural function that fulfills the criterion.
Secondly, due to insufficient prior knowledge, it is challenging to determine the key distance intervals requiring greater discriminability.
Thirdly, because the importance calculation covers all input sentences, the computational complexity of the designed function should be acceptable.
We will explore how to address these challenges in the next section.

\subsection{Distance-importance Function}\label{sec_3_2}
As shown in Figure~\ref{Fig_Framework}, given a sentence-aspect pair $X-Y$, we utilize the Stanza parser\footnote{https://stanfordnlp.github.io/stanza/} developed by the Stanford NLP Group to perform syntactic analysis on $X$ and generate its dependency tree $G$.
The tree $G$ is a special kind of graph whose initial topology encodes different types of directed dependencies across token nodes.
In particular, we regard dependencies as undirected edges, so any two tokens in a sentence are reachable and the minimum tree distance is the number of edges on the shortest path.
Furthermore, the dependency tree $G$ is converted into a settled induced syntactic graph $G_{dis}$, which is fully connected and can be represented as a symmetric adjacency matrix $\mathbf{A}_{dis}\in\mathbb{R}^{N\times N}$, whose entry $\mathbf{A}_{dis}[i,j]$ represents the distance value between the $i$-th and $j$-th tokens.
Because edge weights are usually inversely proportional to their corresponding distance values, we need a distance-importance calculation function to transform $\mathbf{A}_{dis}$.

\begin{figure}[t]
\centering
\includegraphics[width=\columnwidth]{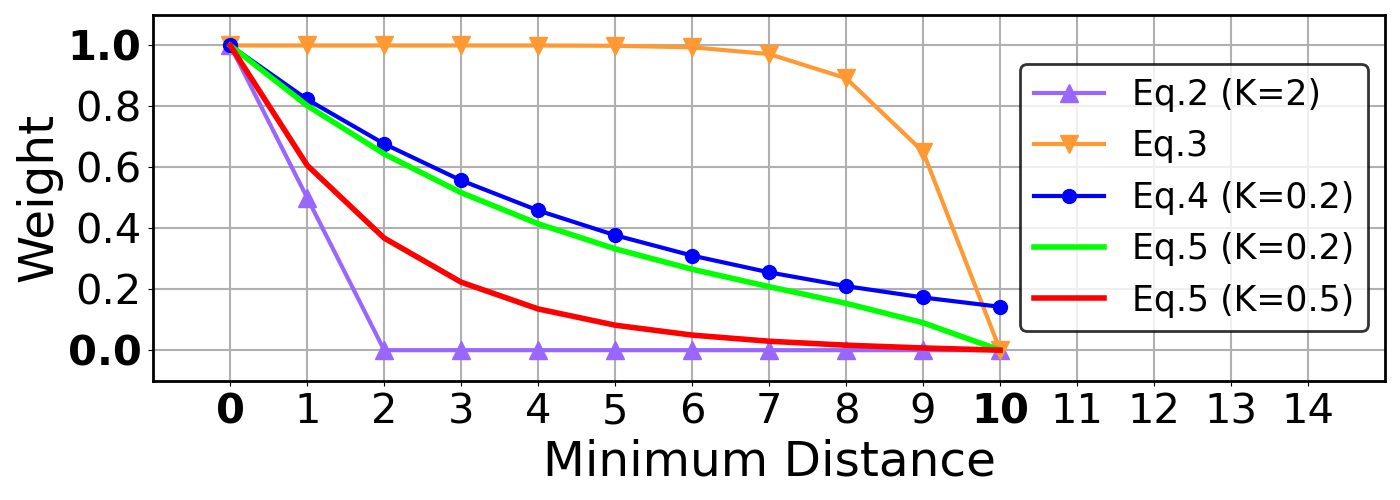}
\vspace{-8mm}
\caption{
Weight distributions for distance-importance functions with different slopes or curvatures.
The boundary value $T$ of the minimum tree distances is 10.
}
\label{Fig_Functions}
\end{figure}

Following the criterion set forth in Section \ref{sec_3_1}, here we design the function $\text{IMP}_{dis}(\cdot)$, which can address three challenges step-by-step.
Specifically, inspired by previous linear functions, a straightforward solution to meet both requirements is to append a constant function to a linear one, which can be expressed as:
\begin{align}
	\text{IMP}_{dis}(t)=\begin{cases}
	1-t/K, &0\leqslant t< K\\
	0, &K\leqslant t\leqslant T\end{cases},
	\label{Eq_2}
\end{align}
where $t$ represents the minimum distance value to be calculated, $T$ represents the predefined distance boundary, and $K$ represents the slope of the linear function.
The function approximately satisfies the criterion, whose importance weight distribution is shown in Figure \ref{Fig_Functions}.
However, such a way of placing different functions in different distance intervals and splicing them is too imprecise, leading to too steep weight changes near the $K$ value.
It is unreasonable that the edge with a distance value of $(K-1)$ has importance, but the next edge with a value of $K$ suddenly has no effect.
In addition, it is hard to determine the optimal $K$ for deciding the key interval.

In order to further enhance $\text{IMP}_{dis}(\cdot)$, here we redesign it from the two requirements of the criterion and replace the above strategy of concatenating functions by intervals.
In particular, we design two sub-functions spanning the entire distance interval $[1,T]$ to satisfy the discriminability and distribution.
To increase discriminability, a power-based sub-function is proposed to maximize the gap of the edge weights of the minimum and maximum distances, which can be expressed as follows:
\begin{align}
	\text{IMP}_{d-1}(t)=1-(t/T)^{T}.
	\label{Eq_3}
\end{align}
To avoid arithmetic and non-smooth distributions, we introduce another exponential-based sub-function to disproportionately reduce the importance of induced dependencies, which can be defined as:
\begin{align}
	\text{IMP}_{d-2}(t)=E^{(-Kt)},
	\label{Eq_4}
\end{align}
where $E$ denotes the natural constant (a.k.a. Euler number), and $K$ is the exponential curvature.
Then, we combine the two sub-functions by a multiplication operation $\oplus$, which can be expressed as follows:
\begin{align}
	\text{IMP}_{dis}(t) = \text{IMP}_{d-1}(t)\oplus \text{IMP}_{d-2}(t)=(1-(t/T)^{T})E^{(-Kt)}.
	\label{Eq_5}
\end{align}
The functions corresponding to Equations (\ref{Eq_2}-\ref{Eq_5}) are shown in Figure \ref{Fig_Functions}.
Since the function $\text{IMP}_{d-2}(\cdot)$ (in blue) controlling the distribution has a maximum weight of 1 while the minimum weight (when $t=T$) is not necessarily very small, the $\text{IMP}_{d-1}(\cdot)$ (in orange) guiding the range of weights simply scales the minimum weight to 0.
In addition, the weight distribution of $\text{IMP}_{d-1}(\cdot)$ is close to 1 in a broad distance interval, protecting the concave arc distribution of $\text{IMP}_{d-2}(\cdot)$.
This is why the non-equidistant distributions of $\text{IMP}_{dis}(\cdot)$ (in green) and $\text{IMP}_{d-2}(\cdot)$ are close to fit.
Compared with the strategy (in purple) of concatenating functions by intervals, the new $\text{IMP}_{dis}(\cdot)$ promotes a smoother distribution while satisfying the criterion, preserving the possibility of exploiting induced dependencies with large distances.

As shown in Figure \ref{Fig_Functions}, different curvatures contribute to different weight distributions and key distance intervals of $\text{IMP}_{dis}(\cdot)$ (red \& green).
Therefore, the selection of the optimal curvature $K$ is crucial, which directly affects the performance of GNN-based ABSA.
From an application point of view, due to insufficient prior knowledge, it is laborious and inefficient to find the optimal $K$ by manual tuning, especially when the candidate set is large.
From an implementation perspective, since the curvature $K$ does not directly participate in model training, it is infeasible to optimize $K$ using backpropagation.
Hence, we use reinforcement learning (RL) \cite{peng2021reinforced,zhao2022multi,zhao2022deep,peng2022reinforced} to search for optimal curvatures for different tasks.
Concretely, we define the problem of finding optimal curvatures as a Two-Armed Bandit \cite{vermorel2005multi} $\{\{a^{+},a^{-}\}, \text{REW}(\cdot), \text{TER}(\cdot)\}$. 
$a^{+}$ and $a^{-}$ denote two actions, $\text{REW}(\cdot)$ is the reward function, and $\text{TER}(\cdot)$ is the termination function:

$\bullet$ \textbf{Action:} 
The action space represents how the RL-based module updates the curvature $K$ according to the reward.
Here, we designate $a^{+}$ and $a^{-}$ as increasing and decreasing a fixed value $S$ to the current $K$ according to the polarity of the reward.

$\bullet$ \textbf{Reward:}
Since we aim to improve ABSA, the gap in validation accuracy by adjacent time intervals is considered a reward indicator.
The reward function can be expressed as follows:
\begin{align}
	\text{REW}(b)=\begin{cases}
	+1,&\text{ACC}(\{X\}_{val},b)> \text{ACC}(\{X\}_{val},b-1)\\
	-1,&\text{ACC}(\{X\}_{val},b)\leqslant \text{ACC}(\{X\}_{val},b-1)\end{cases},
	\label{Eq_6}
\end{align}
where $b$ represents the index of the time interval containing the predefined number of batches, which also implies the update frequency of $K$.
In addition, $\{X\}_{val}$ represents the validation set, $\text{ACC}(\cdot)$ is the function applied to acquire the accuracy of sentiment classification.
Since the key distance interval will gradually shrink as $K$ increases, we perform action $a^{+}$ when the reward is $+1$ (or $a^{-}$ on the contrary), thereby gradually condensing the syntactic information.

$\bullet$ \textbf{Termination:} 
$K$ will be updated continuously until it satisfies:
\begin{align}
	\text{TER}(b)=\left |  {\textstyle \sum_{b-R}^{b}} \text{REW}(b) \right | \leqslant 1,
	\label{Eq_7}
\end{align}
where $R$ is the number of historical rewards.
The inequality suggests that the reward has converged, and $K$ continues constant.
Therefore, the discriminative degree of critical distance intervals is determined dynamically, making RDGCN highly portable across different tasks.

The computational complexity mainly comes from the functions themselves and RL.
The functions cost $O(1+\log(T))$, the RL module costs $O(R+1)$.
Compared with those linear functions, the complexity of the presented $\text{IMP}_{dis}(\cdot)$ changes from a constant level to a linear level, which is still relatively excellent.

\subsection{Type-importance Function}\label{sec_3_3}
Unlike distances, the importance of dependency types is less explicit in the absence of expert knowledge.
Hence, we use a global attention mechanism to calculate the importance weights of type edges.
More specifically, we first number the types and transform the tree $G$ into another induced syntactic graph, $G_{type}$, whose adjacency matrix is $\mathbf{A}_{type}\in\mathbb{R}^{N\times N}$.
Since token nodes usually do not have type-labeled self-loops, we then use custom {\it ``root''} and {\it ``none''} type numbers to fill the diagonal and meaningless induced edges of $\mathbf{A}_{type}$, respectively.
Finally, we initialize the type feature matrix $\mathbf{H}\in\mathbb{R}^{U\times D}$, where $U$ is the total number of all dependency types.
Thus the type-importance function $\text{IMP}_{type}$ can be expressed as follows: 
\begin{equation}
\begin{aligned}
	&\text{IMP}_{type}(u)=
    \begin{cases}
        p[u], &\mathbf{A}[i,j]>0\\ 
        0, &\mathbf{A}[i,j]=0
    \end{cases},\\
    s.t.\ &p=\text{softmax}(\mathbf{H}q)\ \&\ \mathbf{A}_{type}[i,j]=u ,
	\label{Eq_8}
\end{aligned}
\end{equation}
where $q\in\mathbb{R}^{D\times 1}$ is the transposed query vector, and $p\in\mathbb{R}^{1\times U}$ is the weight vector normalized by $\text{softmax}$.
In particular, to preserve the raw topology of the dependency tree $G$, we use its initial adjacency matrix $\mathbf{A}$ as the mask mechanism to remove the induced {\it ``none''} type edges.
In this way, the three syntactic dependency views mentioned in Section \ref{sec_1} are all introduced into RDGCN.

Furthermore, we merge the induced graphs $G_{dis}$ and $G_{type}$ into one graph $G^{*}$ with an addition operation $\oplus$, which can be abstracted into a matrix form as follows:
\begin{equation}
\centering
\begin{aligned}
	\mathbf{A}^{*}=\mathbf{A}_{dis}^{*}&\oplus\mathbf{A}_{type}^{*},\\
    s.t.\ \mathbf{A}_{dis}^{*}=\text{IMP}_{dis}(\mathbf{A}_{dis})&\ \&\ \mathbf{A}_{type}^{*}=\text{IMP}_{type}(\mathbf{A}_{type}),
	\label{Eq_9}
\end{aligned}
\end{equation}
where $\mathbf{A}^{*}$ symbolizes the adjacency matrix of the induced syntactic graph $G^{*}$. 
Each entry $\mathbf{A}^{*}[i,j]$ encodes an induced edge weight and belongs to $[0,2]$.

\subsection{Feature Aggregation}\label{sec_3_4}
In this part, we perform feature aggregation on the adjacency matrix $\mathbf{A}^{*}\in\mathbb{R}^{N\times N}$ and the initial feature matrix $\mathbf{E}=\mathbf{E^{0}}\in\mathbb{R}^{N\times D}$ obtained by the sentence encoder to enhance the token representations.
Since $\mathbf{A}^{*}$ already has the weights, we implement the aggregation function $\text{AGG}(\cdot)$ by convolution aggregation, whose aggregation process of the $l$-th layer can be expressed as:
\begin{align}
	\mathbf{E}^{l}=\sigma(\mathbf{A}^{*}\mathbf{E}^{(l-1)}\mathbf{W}^{l}),
	\label{Eq_10}
\end{align}
where $\mathbf{W}^{l}$ represents the feature transformation matrix of the $l$-th layer.
After iterating $L$ times, we obtain the final feature representation matrix $\mathbf{E}^{L}\in\mathbb{R}^{N\times D}$ of the sentence $X$.

\subsection{Pooling and Classification}\label{sec_3_5}
Since the aspect $Y=<$$y_1, y_2, ..., y_M$$>$ is a sub-sequence of $X$, we first filter out the non-aspect features of $\mathbf{E}^{L}$ to obtain the aspect-specific feature matrix $\mathbf{F}\in\mathbb{R}^{M\times D}$, and then perform mean pooling on $\mathbf{F}$ to obtain the aspect-specific vector, which can be expressed as follows:
\begin{align}
	f=(f_{1}+\cdots+f_{M})/M,\ s.t.\ f_{i}=\mathbf{F}[i]\in\mathbb{R}^{1\times D}\ \&\ \mathbf{F}=\mathbf{E}^{L}[Y].
	\label{Eq_11}
\end{align}
Then, we feed $f$ into a classifier consisting of a linear function and a $\text{softmax}$ to yield a probability distribution over the polarity decision.
Finally, we optimize the parameters based on the cross-entropy loss:
\begin{align}
	\mathcal{L}=-\sum_{\{X\}_{train}}\text{log}(\text{softmax}(f^{*}\mathbf{Z}+b))c^{*},
	\label{Eq_12}
\end{align}
where $\{X\}_{train}$ is a set containing all training $X-Y$ pair samples, $f^{*}$ is the $D$-dimensional vector of each training aspect, $\mathbf{Z}\in\mathbb{R}^{D\times C}$ and $b\in\mathbb{R}^{1\times C}$ indicate the trainable parameters and bias of the classifier.
$c^{*}\in\mathbb{R}^{C\times 1}$ is the transposed polarity label vector corresponding to $f^{*}$, and $C$ denotes the total number of polarity categories.
\section{Experiments}
\label{Sec_4}
In this section, we present the experimental settings consisting of datasets and evaluation, baselines, and implementation details.
We then perform classification tasks, case study, ablation study, etc., to address three research questions (RQs):
\begin{itemize}[leftmargin=*]
\item \textbf{RQ1}: How does RDGCN perform on the ABSA dataset compared to state-of-the-art (SOTA) baselines? 
\item \textbf{RQ2}: How much do the syntactic importance functions included in RDGCN improve performance? 
\item \textbf{RQ3}: How much does changing important hyperparameters of RDGCN affect ABSA? 
\end{itemize}

\subsection{Datasets and Evaluation}\label{Sec_4_1}
Following previous ABSA works, we evaluate the proposed RDGCN on three popular fine-grained datasets, namely Restaurant, Laptop, and Twitter.
Among them, Restaurant and Laptop are from SemEval-2014 task 4 \cite{Pontiki2014SemEval}, which comprise sentiment reviews from restaurant and laptop domains, respectively.
Moreover, Twitter is collected and processed by \cite{dong2014adaptive} from tweets.
Following most studies like \cite{chen2017recurrent,li2021dual,dai2021does}, we remove these samples with conﬂicting polarities or with {\it ``NULL''} aspects in all datasets, where each aspect is annotated with one of three polarities: positive, negative, and neutral.
In order to measure the effectiveness of all methods, we utilize two metrics, i.e., accuracy (Acc.) and macro-F1 (F1), to expose their classification performance.

\renewcommand{\arraystretch}{0.8}
\begin{table*}[t]
\centering
\caption{
Classification results (\%).
The best results for all models are in bold, while the second-best results are in italics.
}
\vspace{-2.6mm}
\label{Tab_3}
\resizebox{\textwidth}{!}{
\begin{tabular}{c|cccccccc}
\hline
\small
\multirow{2}{*}{Model} & \multirow{2}{*}{Type} & \multirow{2}{*}{Syntactic View} & \multicolumn{2}{c}{Restaurant} & \multicolumn{2}{c}{Laptop} & \multicolumn{2}{c}{Twitter} \\ \cline{4-9} 
 &  &  & Acc. & F1 & Acc. & F1 & Acc. & F1 \\ \hline
ATAE-LSTM~\cite{wang2016attention} & RNN+Attention & - & 77.20 & - & 68.70 & - & - & - \\
IAN~\cite{ma2017interactive} & RNN+Attention & - & 78.60 & - & 72.10 & - & - & - \\
RAM~\cite{chen2017recurrent} & RNN+Attention & - & 80.23 & 70.80 & 74.49 & 71.35 & 69.36 & 67.30 \\
MGAN~\cite{fan2018multi} & RNN+Attention & - & 81.25 & 71.94 & 75.39 & 72.47 & 72.54 & 70.81 \\ \hline
TNet~\cite{li2018transformation} & RNN+CNN & - & 80.69 & 71.27 & 76.54 & 71.75 & 74.90 & 73.60 \\ 
PWCN~\cite{zhang2019syntax} & RNN+Syntax & Distance & 80.96 & 72.21 & 76.12 & 72.12 & - & - \\ \hline
ASGCN~\cite{zhang2019aspect} & RNN+GCN+Syntax & Topology & 80.77 & 72.02 & 75.55 & 71.05 & 72.15 & 70.40 \\
TD-GAT~\cite{huang2019syntax} & RNN+GAT+Syntax & Topology & 81.20 & - & 74.00 & - & - & - \\
BiGCN~\cite{zhang2020convolution} & RNN+GCN+Syntax & Topology \& Type & 81.97 & 73.48 & 74.59 & 71.84 & 74.16 & 73.35 \\
kumaGCN~\cite{chen2020inducing} & RNN+GCN+Syntax+Semantics & Topology & 81.43 & 73.64 & 76.12 & 72.42 & 72.45 & 70.77 \\
DGEDT~\cite{tang2020dependency} & RNN+GCN+Syntax & Topology & 83.90 & 75.10 & 76.80 & 72.30 & 74.80 & 73.40 \\
R-GAT~\cite{wang2020relational} & RNN+GAT+Syntax & Topology \& Type \& Distance & 83.30 & 76.08 & 77.42 & 73.76 & 75.57 & 73.82 \\
DualGCN~\cite{li2021dual} & RNN+GCN+Syntax+Semantics & Topology & 84.27 & \textbf{78.08} & 78.48 & 74.74 & 75.92 & 74.29 \\
SSEGCN~\cite{zhang2022ssegcn} & RNN+GCN+Syntax+Semantics & Topology \& Distance & \textbf{84.72} & 77.51 & \textit{79.43} & \textit{76.49} & \textit{76.51} & \textit{75.32} \\
$\text{RDGCN}^{*}$ & RNN+GCN+Syntax & Topology \& Type \& Distance & \textit{84.36} & \textit{78.06} & \textbf{79.59} & \textbf{76.75} & \textbf{76.66} & \textbf{75.37} \\ 
\hline
BERT~\cite{kenton2019bert} & PLM & - & 85.97 & 80.09 & 79.91 & 76.00 & 75.92 & 75.18 \\
DGEDT+BERT~\cite{tang2020dependency} & PLM+GCN+Syntax & Topology & 86.30 & 80.00 & 79.80 & 75.60 & \textit{77.90} & 75.40 \\
R-GAT+BERT~\cite{wang2020relational} & PLM+GAT+Syntax & Topology \& Type \& Distance & 86.60 & \textbf{81.35} & 78.21 & 74.07 & 76.15 & 74.88 \\
T-GCN+BERT~\cite{tian2021aspect} & PLM+GCN+Syntax & Topology \& Type & 86.16 & 79.95 & 80.88 & 77.03 & 76.45 & 75.25 \\
DualGCN+BERT~\cite{li2021dual} & PLM+GCN+Syntax+Semantics & Topology & 87.13 & 81.16 & \textit{81.80} & \textit{78.10} & 77.40 & 76.02 \\
SSEGCN+BERT~\cite{zhang2022ssegcn} & PLM+GCN+Syntax+Semantics & Topology \& Distance & \textit{87.31} & 81.09 & 81.01 & 77.96 & 77.40 & \textit{76.02} \\
$\text{RDGCN+BERT}^{*}$ & PLM+GCN+Syntax & Topology \& Type \& Distance & \textbf{87.49} & \textit{81.16} & \textbf{82.12} & \textbf{78.34} & \textbf{78.29} & \textbf{77.14} \\ \hline
\end{tabular}}
\end{table*}

\subsection{Baselines}\label{Sec_4_2}
To comprehensively evaluate the performance of RDGCN, we compare it with SOTA baselines, which are briefly described as follows:

\noindent 1) \textbf{ATAE-LSTM} \cite{wang2016attention} is an attention-based LSTM model that focuses on aspect-specific key parts of sentences.

\noindent 2) \textbf{IAN} \cite{ma2017interactive} interactively calculates attention scores for aspects and contexts, yielding aspect and context representations, respectively.

\noindent 3) \textbf{RAM} \cite{chen2017recurrent} leverages a recurrent attention mechanism on sentence memory to extract aspect-specific importance information.

\noindent 4) \textbf{MGAN} \cite{fan2018multi} applies a fine-grained attention mechanism to capture token-level interactions between aspects and contexts.

\noindent 5) \textbf{TNet} \cite{li2018transformation} transforms token representations from a BiLSTM into target-specific representations and then uses the CNN layer instead of attention to generate salient features for sentiment classification.

\noindent 6) \textbf{PWCN} \cite{zhang2019syntax} calculates the proximity weights of context words for the aspect according to the minimum distances over the dependency tree, and applies these weights to enhance the output of BiLSTM to obtain aspect-specific syntax-aware representations. 

\noindent 7) \textbf{ASGCN} \cite{zhang2019aspect} applies GCN on the raw topology of the dependency tree to introduce syntactic information.

\noindent 8) \textbf{TD-GAT} \cite{huang2019syntax} leverages a graph attention network (GAT) \cite{velivckovic2018graph} to capture syntactic dependency structures.

\noindent 9) \textbf{BiGCN} \cite{zhang2020convolution} performs convolutions over hierarchical lexical and syntactic graphs to integrate token co-occurrence information and dependency type information.

\noindent 10) \textbf{kumaGCN} \cite{chen2020inducing} associates dependency trees with aspect-specific induced graphs, and applies gating mechanisms to obtain syntactic features with latent semantic information.

\noindent 11) \textbf{DGEDT} \cite{tang2020dependency} jointly considers the representations learned from a Transformer and graph-based representations learned from the corresponding dependency graph in an iterative interactive manner.

\noindent 12) \textbf{R-GAT} \cite{wang2020relational} transforms the dependency tree into a star-induced graph with edges consisting of minimum distances and dependency types, and introduces a relational GAT for aggregation via attention.

\noindent 13) \textbf{T-GCN} \cite{tian2021aspect} distinguishes relation types via attention, and uses an attentive layer ensemble to learn features from many GCN layers.

\noindent 14) \textbf{DualGCN} \cite{li2021dual} simultaneously introduces syntactic information and semantic information through SynGCN and SemGCN modules.

\noindent 15) \textbf{SSEGCN} \cite{zhang2022ssegcn} acquires semantic information through attention, and equips it with syntactic information of minimum tree distances.

\noindent 16) \textbf{BERT \& Model+BERT} \cite{kenton2019bert} represent the pre-trained language model (PLM) BERT and the model with BERT as a sentence encoder.

\subsection{Implementation Details}\label{Sec_4_3}
For all experiments, we employ pre-trained 300-dimensional Glove vectors \cite{pennington2014glove} to initialize token embeddings.
Following the common settings of previous studies such as \cite{li2021dual,zhang2022ssegcn}, we vectorize the part-of-speech (POS) information of tokens and their relative position with respect to the boundary tokens of the aspect.
Then, we concatenate the 30-dimensional POS and position vectors with the Glove vectors, and input them into a BiLSTM model to get the initial token feature representations.
In addition, we set the hidden dimension of BiLSTM and GCN to $D=50$, the number of model layers to $L=2$.
To ensure the optimization space, we leverage the dropout of 0.7 and 0.1 to the input of BiLSTM and the output of the framework (BiLSTM \& GCN), respectively.
We optimize RDGCN\footnote{https://github.com/RDGCN/RDGCN} using the Adam optimizer with a learning rate of 0.002, where the total number of training epochs is 20 and the batch size is 32.
For the introduction of syntax, we utilize an off-the-shelf Stanza parser to get syntactic dependency trees, and compute the minimum tree distances among tokens (including inner tokens of the aspect term).
The upper bound of the distance values is set to $T=10$.
Since the datasets do not contain the validation set, the test accuracy is used to implement the reward function $\text{ACC}(\cdot)$.
In addition, we set the predefined range of curvature to $K\in[0.1,2]$, the update value for $K$ of RL actions is $S=0.1$, the update frequency is $2$ (i.e., $b$ contains 2 batches), the size is $R=10$, $K$ is initialized to 0.1 because the initial rewards are all +1 as performance increases.
For \textbf{Model+BERT}, we utilize the bert-base-uncased\footnote{https://github.com/huggingface/transformers} English version.

\renewcommand{\arraystretch}{1.2}
\begin{table}[t]
\centering
\caption{
Case study results.
Different aspects within the same sentence are colored differently.
The P, N, and O are positive, negative, and neutral, respectively.
The Label column merges the same RDGCN column as it, which displays the true labels.
}
\vspace{-2.6mm}
\label{Tab_4}
\resizebox{\columnwidth}{!}{
\begin{tabular}{c|ccc}
\hline
Sentence & R-GAT & SSEGCN & \begin{tabular}[c]{@{}c@{}}RDGCN\\ (Label)\end{tabular} \\ \hline
Great \textbf{\textcolor{red}{food}} but the \textbf{\textcolor{blue}{service}} was dreadful! & N-N & P-N & P-N \\ \hline
\begin{tabular}[c]{@{}c@{}}Can you buy any laptop that\\ matches the \textbf{\textcolor{red}{quality}} of a MacBook?\end{tabular} & P & O & P \\ \hline
Biggest complaint is \textbf{\textcolor{red}{Windows 8}}. & O & N & N \\ \hline
Try the \textbf{\textcolor{red}{rose roll}} (not on \textbf{\textcolor{blue}{menu}}). & P-N & P-O & P-O \\ \hline
\end{tabular}}
\end{table}

\begin{figure*}[t]
\centering
\subfigure{
    \begin{minipage}[t]{0.24\textwidth}
        \includegraphics[width=\textwidth]{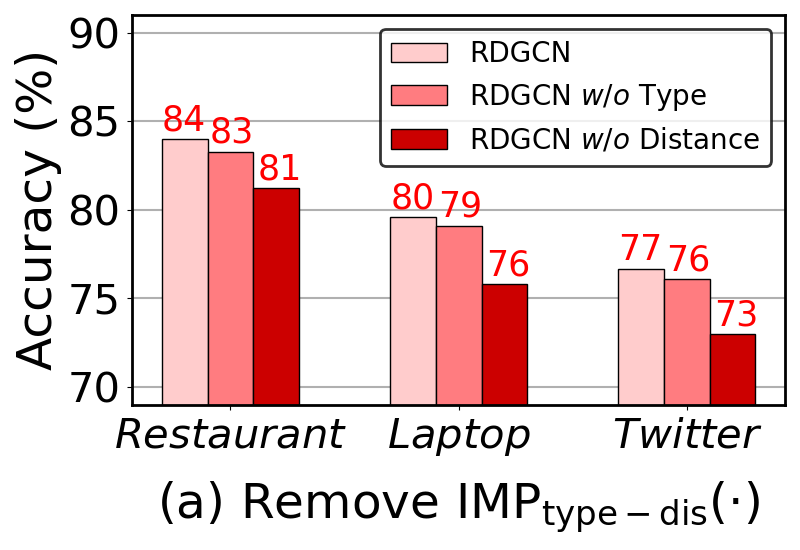}
    \end{minipage}}
\subfigure{
    \begin{minipage}[t]{0.24\textwidth}
        \includegraphics[width=\textwidth]{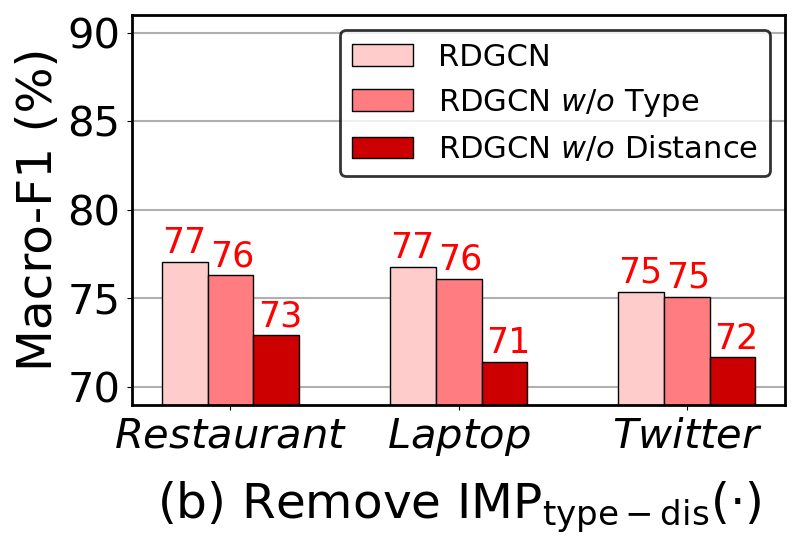}
    \end{minipage}}
\subfigure{
    \begin{minipage}[t]{0.24\textwidth}
        \includegraphics[width=\textwidth]{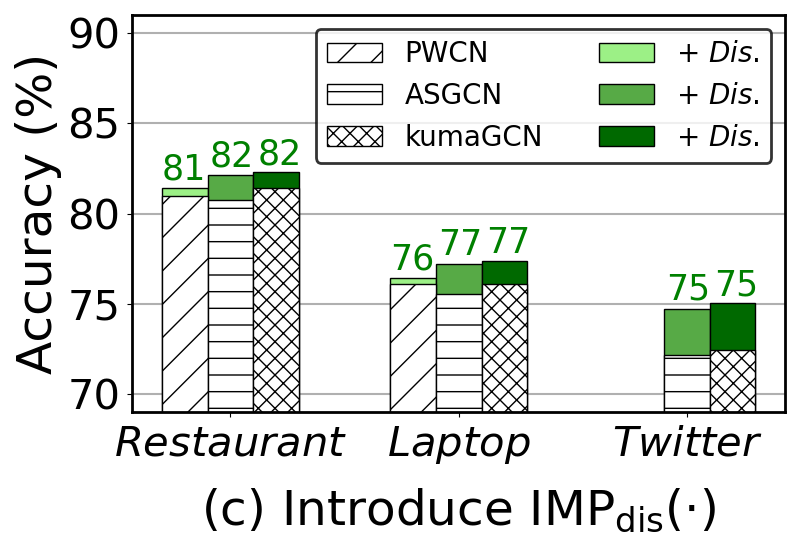}
    \end{minipage}}
\subfigure{
    \begin{minipage}[t]{0.24\textwidth}
        \includegraphics[width=\textwidth]{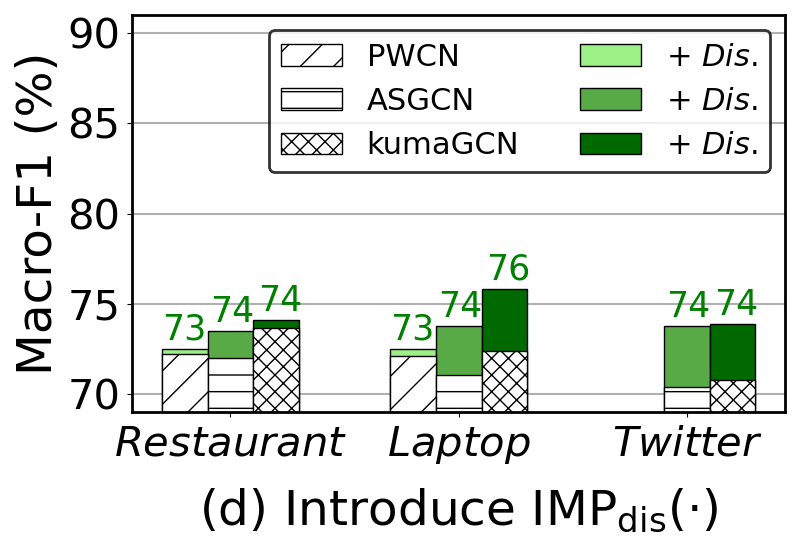}
    \end{minipage}}
\vspace{-5.5mm}
\caption{
Visualization of ablation study results, where {\it ``w/o''} is without and {\it ``+ Dis.''} means the introduction of the distance syntax.
The left two figures depict the removal of the importance function in RDGCN, while the right two show the introduction of the distance-importance function $\text{IMP}_{dis}(\cdot)$ to the baselines, where the green bars indicate the improved performance.
}
\label{Fig_Ablation}
\end{figure*}

\subsection{Classification Results}\label{Sec_4_4}
To answer \textbf{RQ1}, we compare our RDGCN with all baselines on three ABSA datasets, and the classification results are depicted in Table \ref{Tab_3}.
The classification results justify that the proposed RDGCN (+BERT) exhibits overall better sentiment performance than SOTA baselines.
However, RDGCN lags behind DualGCN as well as SSEGCN on the Restaurant dataset.
A possible explanation is that the two baselines additionally capture the semantic information of sentences based on self-attention, which refines the aspect representations.
Because the features encoded by BERT already contain rich semantics, RDGCN-BERT remedies this deficiency in the accuracy metric.
Moreover, we can observe that the performance of RNN+attention-based baselines is generally weaker than that of RNN+GNN-based baselines.
This is because GNN enhances aspect representations by learning syntactic dependency trees or induced trees, which shows that capturing the structural patterns of syntactic parsing can indeed improve analysis performance.
Among GNN-based ones, models (such as R-GAT and SSEGCN) that consider multiple syntactic views outperform models, such as DGEDT and DualGCN, that only consider the raw topology of dependency trees, particularly on the Laptop and Twitter datasets.
This implies that syntaxes in different views may be complementary, and effectively and comprehensively mining syntactic information can further improve GNN-based ABSA.
Further, the existing SOTA baselines that incorporate distance syntax either intuitively reduce the distance weights equidistantly (PWCN) or obstruct the original explicit weights by attention and masks (R-GAT and SSEGCN), both of which are inferior to RDGCN.
It shows that the proposed criterion together with the distance-importance function are necessary and effective.
Last but not least, we can observe that the powerful BERT outperforms most baselines and RDGCN+BERT achieves the biggest breakthrough in BERT performance compared to other algorithms, justifying that RDGCN acquires more valuable syntactic knowledge for ABSA.
In general, RDGCN (+BERT) performs the best on ABSA tasks compared to the SOTA baselines.

\subsection{Case Study}\label{Sec_4_5}
To better showcase the superiority of our RDGCN, we conduct case studies on some example sentences, as depicted in Table \ref{Tab_4}.
The first sentence {\it ``Great food but the service was dreadful''} owns two aspects ({\it ``food’‘} and {\it ``service’‘}) with opposite sentiment polarities.
The aspect {\it ``quality’‘} in the second sentence does not have any obvious opinion token.
The interfering token {\it ``Biggest’‘} in the third sentence {\it ``Biggest complaint is Windows 8''} may neutralize the negativity of the opinion token {\it ``complaint''}.
The fourth sample has the above three difficulties at the same time.
On the one hand, we argue that the estimations of the attention-based R-GAT are susceptible to opposite or interfering opinion tokens in the first, third, and fourth sentences.
On the other hand, SSEGCN fails to deal with the second sentence lacking explicit opinion tokens.
A possible explanation is that the dependency type syntax is more suitable for such cases than the tree distance syntax, which is not available in SSEGCN.
Consistent predictions with true labels verify that RDGCN captures more complementary syntactic information than SOTA baselines.

\begin{figure}[t]
\centering
    \includegraphics[width=0.495\columnwidth]{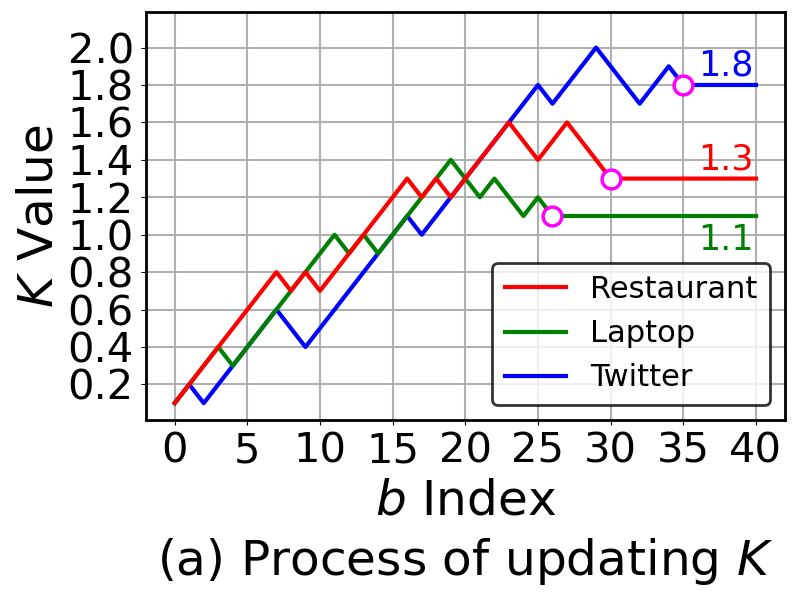}
    \includegraphics[width=0.495\columnwidth]{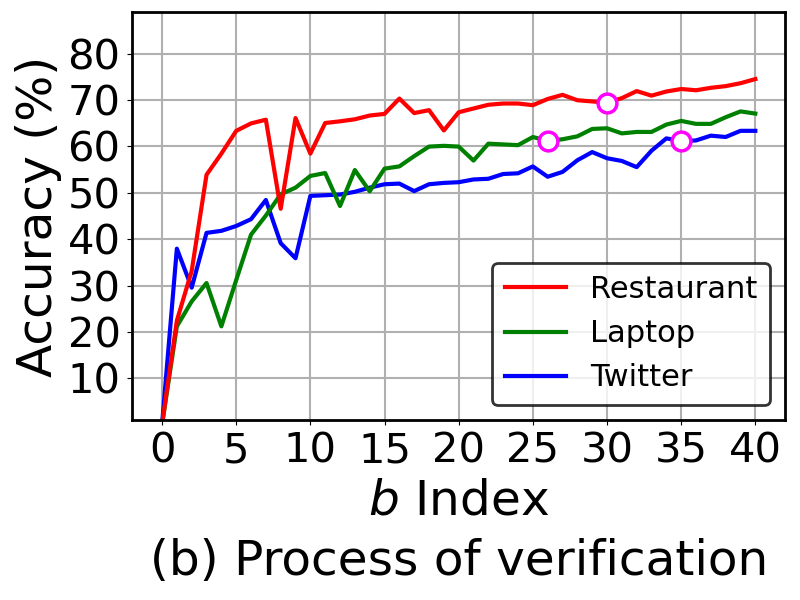}
\vspace{-7.5mm}
\caption{
The process of the RL module, whose hollow points indicate that RL stops searching at the current time index $b$.
}
\label{Fig_RL_1}
\vspace{-2.5mm}
\end{figure}

\begin{figure}[t]
\centering
    \includegraphics[width=0.495\columnwidth]{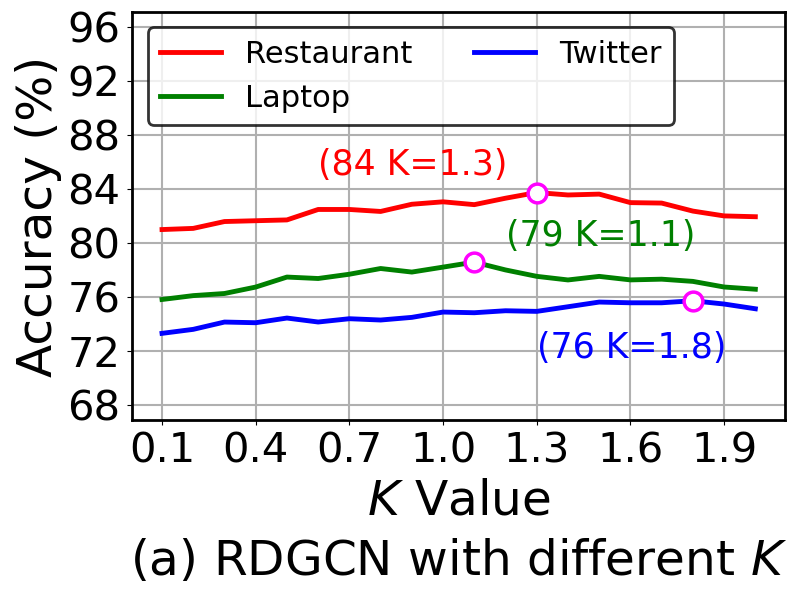}
    \includegraphics[width=0.495\columnwidth]{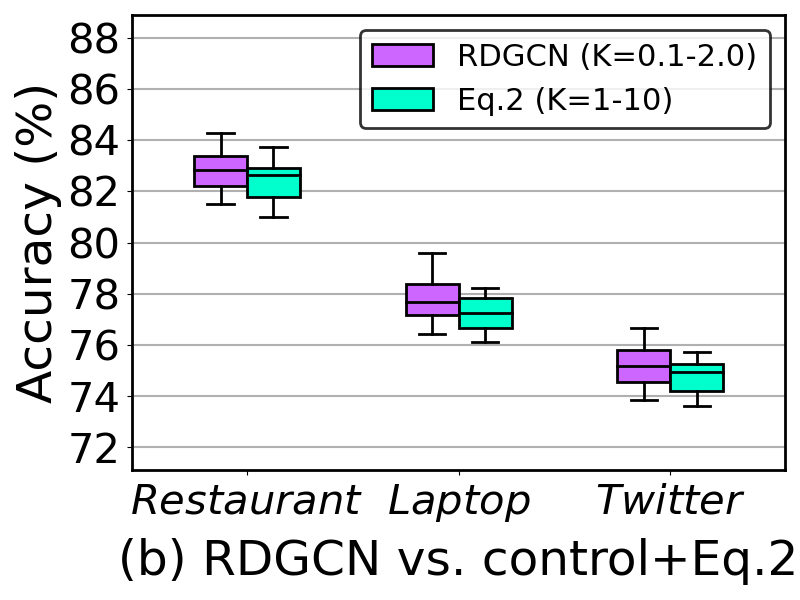}
\vspace{-7.5mm}
\caption{
The effect of curvature values on performance.
The left figure depicts the performance of RDGCN with different $K$, while the right shows the performance difference between RDGCN and a Eq. \ref{Eq_2}-based control.
}
\label{Fig_RL_2}
\end{figure}

\begin{figure*}[t]
\centering
\subfigure{
    \begin{minipage}[t]{0.325\textwidth}
        \includegraphics[width=\textwidth]{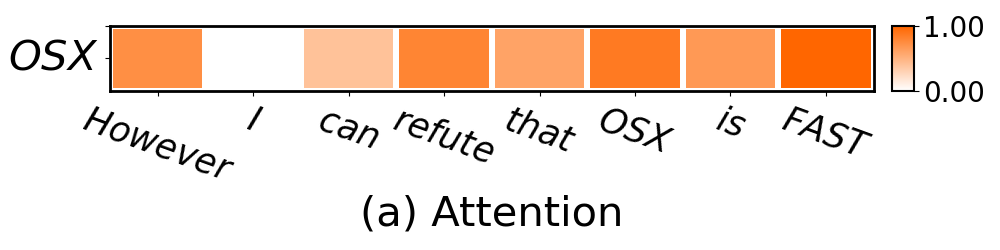}
    \end{minipage}}
\subfigure{
    \begin{minipage}[t]{0.325\textwidth}
        \includegraphics[width=\textwidth]{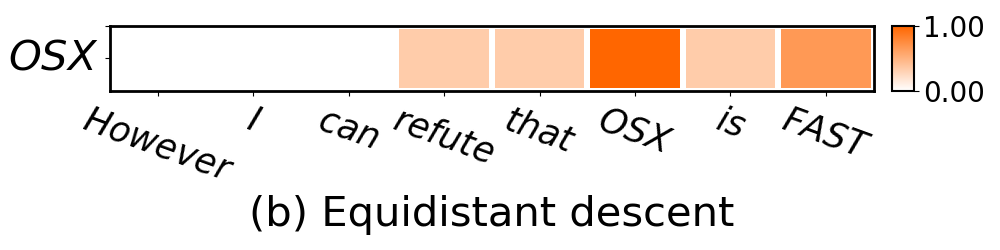}
    \end{minipage}}
\subfigure{
    \begin{minipage}[t]{0.325\textwidth}
        \includegraphics[width=\textwidth]{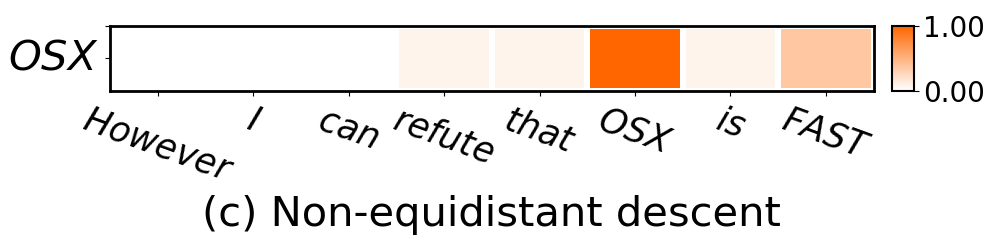}
    \end{minipage}}
\vspace{-8mm}
\caption{
Visualization of $[0,1]$-scaled importance weights of context tokens for the aspect {\it ``OSX''} using different strategies.
}
\label{Fig_RL_3}
\vspace{-1mm}
\end{figure*}

\subsection{Ablation Study}\label{Sec_4_6}
To answer \textbf{RQ2}, we conduct ablation studies to examine the effect of syntactic importance functions on model performance.
As depicted in Figure \ref{Fig_Ablation}(a) and Figure \ref{Fig_Ablation}(b), the performance of RDGCN decreases regardless of whether it is without the distance-importance function $\text{IMP}_{dis}(\cdot)$ (distance syntax) or the type calculation $\text{IMP}_{type}(\cdot)$ (type and topology syntax).
It is worth noting that dropping the function $\text{IMP}_{dis}(\cdot)$ will result in a larger performance penalty than dropping the function $\text{IMP}_{type}(\cdot)$.
RDGCN without $\text{IMP}_{type}(\cdot)$ reduces both the Acc. and F1 by about 1\%, while RDGCN with $\text{IMP}_{dis}(\cdot)$ removed decreases the Acc. by about 3\% and F1 by about 4\% on average for the three datasets.
This observation suggests that the $\text{IMP}_{dis}(\cdot)$ may be more critical to aspect analysis than $\text{IMP}_{type}(\cdot)$.
To further examine the applicability of $\text{IMP}_{dis}(\cdot)$, we graft it to three baselines to replace the original distance-importance calculation procedure (PWCN) or dependency topology (ASGCN and kumaGCN).
From the outcomes illustrated in Figure \ref{Fig_Ablation}(c) and Figure \ref{Fig_Ablation}(d), we draw three conclusions.
First, the performance of all three baselines is improved, suggesting that the investigation of distance syntax and the proposed criterion are warranted.
Second, since PWCN does not apply the $\text{IMP}_{dis}(\cdot)$ to the edge weights of syntactic induced graphs and also does not learn structural patterns through GNNs, its improvement is limited.
Third, the amelioration of syntactic distance-based ASGCN and kumaGCN is more significant on Laptop and Twitter than on Restaurant, which is consistent with the first conclusion about the classification results in Table \ref{Tab_3}, again showing that $\text{IMP}_{dis}(\cdot)$ takes full advantage of the explicit distance syntax and has good portability on several datasets.
Overall, the importance calculation functions contained in RDGCN for the minimum distances and dependency types are efficacious in improving performance, particularly the former, which dynamically searches for exponential curvature based on reinforcement learning (RL) and generates weight distributions that vary with the interval.

\begin{figure}[t]
\centering
\subfigure{
    \begin{minipage}[t]{0.485\columnwidth}
        \includegraphics[width=\textwidth]{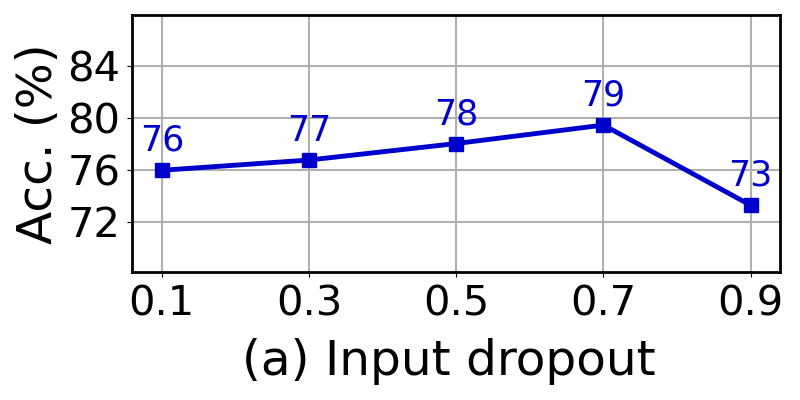}\\
        \includegraphics[width=\textwidth]{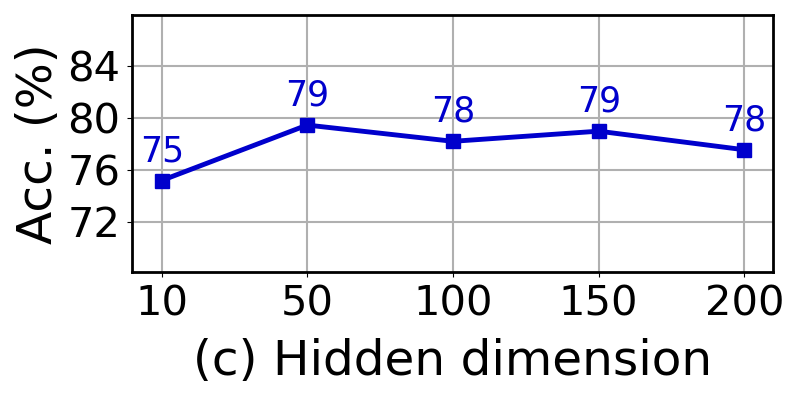}
    \end{minipage}}
\subfigure{
    \begin{minipage}[t]{0.485\columnwidth}
        \includegraphics[width=\textwidth]{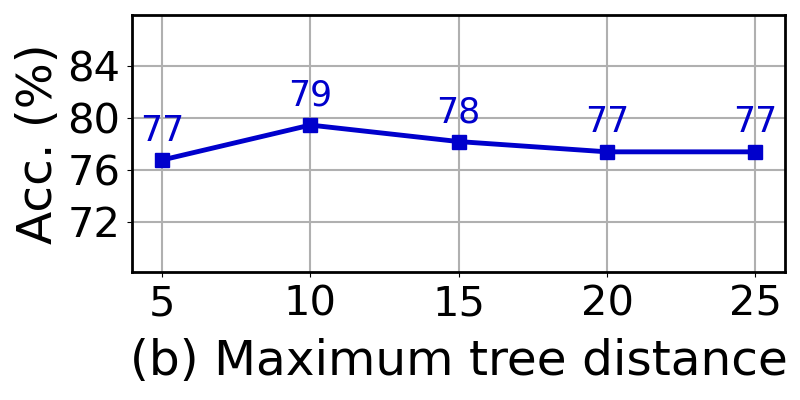}\\
        \includegraphics[width=\textwidth]{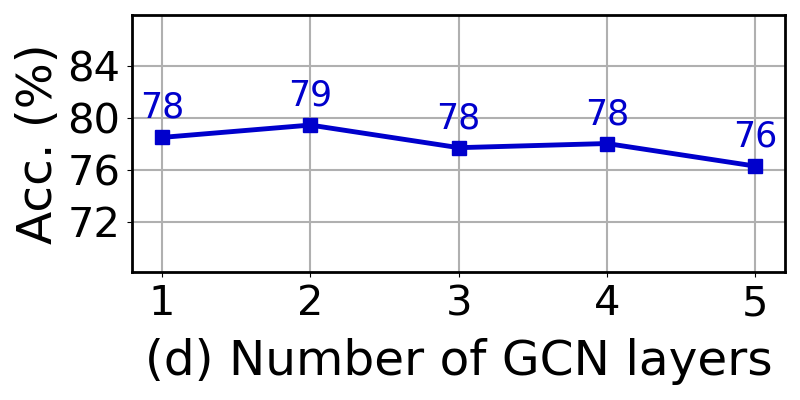}
    \end{minipage}}
\vspace{-6mm}
\caption{
Visualization of the performance impact of different key hyperparameters on the Laptop dataset.
}
\label{Fig_Para}
\end{figure}

\subsection{RL Process Analysis}\label{Sec_4_7}
In this part, we focus on the RL-based module to further examine the distance-importance function $\text{IMP}_{dis}(\cdot)$.
To present the RL process, we plot the updating of the curvature $K$ that controls the concave arc distribution of the distance edge weights in Figure \ref{Fig_RL_1} (a).
As the time index $b$ increases, we can observe that the curvatures corresponding to different datasets are updated towards disparate destinations and remain unchanged after the termination condition is met.
Although $K$ does not directly participate in training, it will influence the edge weights of induced graphs, which are the important factors affecting performance.
Therefore, we tune the history reward number $R$ small to speed up the end of RL.
As per Figure \ref{Fig_RL_1}(b), it is clear that after $K$ is fixed, the performance of RDGCN is still improving.
In this way, the RL module can quickly achieve the best $K$, making the performance improve faster, while ensuring a long-period stable training process.
In other words, because the training is less stable during the update process, RL chooses $K$ during the early period of faster performance improvement, and instructs RDGCN with stable $K$ in the subsequent process.
To evaluate the search results, we quantify the influence of $K$ on model performance in Figure \ref{Fig_RL_2}(b).
On the one hand, different $K$ values yield different performance on every dataset.
This is expected since different $K$ determine different key distance intervals.
On the other hand, the $K$ values searched for different tasks all facilitate the RDGCN to yield excellent performance, indicating the effectiveness of the RL module.
Further, we swap the function $\text{IMP}_{dis}(\cdot)$ with the importance function via interval concatenation adopted in Equation \ref{Eq_2} to construct the control model.
The box plots in Figure \ref{Fig_RL_2}(b) depict the average results yielded from RDGCN and its control baseline on all optional curvatures ($K\in [0.1,2]$) or slopes ($K\in [1,10]$).
We find that the $\text{IMP}_{dis}(\cdot)$-based RDGCN outperforms the control model on all datasets, which implies that our smooth descent strategy is more feasible and reasonable than steep descent.
Besides, we compare the function $\text{IMP}_{dis}(\cdot)$ with attention-based and equidistant strategies on the sentence-aspect pair sample {\it ``However I can refute that OSX is FAST''}-{\it ``OSX''}.
Figure \ref{Fig_RL_3} illustrates the importance weights of context tokens for the aspect term {\it ``OSX''} based on different strategies.
First, even though the attention allocates a greater importance weight to the opinion token {\it ``FAST''}, it pays some attention to noise tokens like {\it ``However''} and {\it ``refute''} that may interfere with classification.
Second, the two types of weight descent also allocate greater importance to the key opinion token {\it ``FAST''} while eliminating the interference of {\it ``However''}.
Third, the distance-importance function $\text{IMP}_{dis}(\cdot)$ using non-equidistant descent further excludes all contexts except {\it ``FAST''} compared to the equidistant descent method, which is beneficial for improving ABSA.
Thus, it makes sense to design a function for the importance calculation of explicit distance syntax via the criterion.

\subsection{Hyperparameter Analysis}\label{Sec_4_8}
To answer \textbf{RQ3}, we examine the influence of four hyperparameters (the input dropout, the maximum of the tree-based distances $T$, the hidden dimension $D$, and the number of GCN layers $L$) of RDGCN on ABSA performance of the Laptop dataset, as depicted in Figure \ref{Fig_Para}.
For the dropout, the performance first increases and then decreases because too small or too large dropout value will lead to overfitting or insufficient input features.
For the $T$, a lower upper bound of the distances may miss part of the useful syntax for ABSA.
In addition, due to the requirements of the proposed criterion on the weights of induced edges with larger distances, the performance does not show a significant landslide as $T$ increases.
For the $D$, feature vectors and matrices with too small dimensions are difficult to encode sufficient feature information, resulting in sub-optimal performance.
For the $L$, too many aggregation layers may cause feature over-smoothing, which is common in GNN-based ABSA models.
Based on the above observations, we argue that the influence of these hyperparameters in the regular range on the aspect sentiment prediction performance of RDGCN is acceptable, which once again verifies that the proposed RDGCN has excellent stability.
\section{Conclusion}
This paper develops RDGCN to improve the importance calculation of dependency types and tree-based distances for ABSA.
Extensive experiments justify the effectiveness of RDGCN.

\section*{Acknowledgements}
This research is supported by National Key R\&D Program of China through grant 2022YFB3104700, NSFC under grant 62322202.
Philip S. Yu was supported in part by NSF under grant III-2106758.

\bibliographystyle{ACM-Reference-Format}
\bibliography{6-Referencce}

\end{document}